\documentclass[10pt, conference, anonymous, compsoc]{IEEEtran}

\ifCLASSOPTIONcompsoc
  \usepackage[nocompress]{cite}
\else
  \usepackage{cite}
\fi

\ifCLASSOPTIONcompsoc
 \usepackage[pdftex]{graphicx}
 \graphicspath{{fig/}}
\else
 \usepackage[dvips]{graphicx}
 \graphicspath{{fig/}}
\fi
\usepackage{graphicx}
\usepackage{textcomp}  
\usepackage{scalerel}  

\def \arm{\includegraphics[scale=0.05]{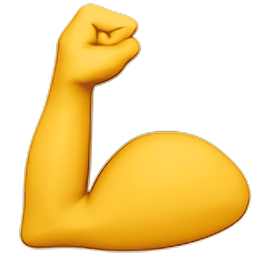}}
\def \sunglass{\includegraphics[scale=0.05]{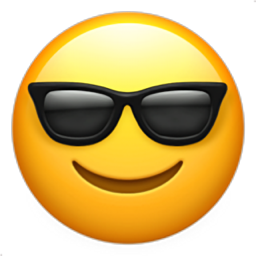}}

\usepackage[skip=1pt]{caption}
\usepackage[ruled,vlined,linesnumbered,noresetcount]{algorithm2e}
\usepackage{amssymb}
\usepackage{amsmath}
\usepackage{bm}

\usepackage{csquotes}

\ifCLASSOPTIONcompsoc
 \usepackage[caption=false,font=footnotesize,labelfont=sf,textfont=sf]{subfig}
\else
 \usepackage[caption=false,font=footnotesize]{subfig}
\fi

\usepackage{fixltx2e}

\usepackage{pdfpages}
\usepackage{hyperref}
\usepackage{cleveref}

\begin{document}

\title{What Sentiment and Fun Facts We Learnt
Before FIFA World Cup Qatar 2022 Using Twitter and AI}



\author{
    \IEEEauthorblockN{
        James She\IEEEauthorrefmark{2}, Kamilla~Swart-Arries\IEEEauthorrefmark{2},  
        Mohammad Belal\IEEEauthorrefmark{2} and Simon~Wong\IEEEauthorrefmark{4} 
    }
    \IEEEauthorblockA{
        \IEEEauthorrefmark{2}College of Science and Engineering, Hamad Bin Khalifa University, Qatar\\
        \IEEEauthorrefmark{4}Department of Electronic and Computer Engineering, HKUST, Hong Kong\\
    }
    \IEEEauthorblockA{
        Email: \IEEEauthorrefmark{2}\{pshe, kswartarries, mobe50543\}@hbku.edu.qa, \IEEEauthorrefmark{4}tywongbf@connect.ust.hk
    }
 }

\maketitle

\IEEEtitleabstractindextext{%
\begin{abstract}
Twitter is a social media platform bridging most
countries and allows real-time news discovery. Since the tweets
on Twitter are usually short and express public feelings, thus
provide a source for opinion mining and sentiment analysis
for global events. This paper proposed an effective solution, in
providing a sentiment on tweets related to the FIFA World
Cup. At least 350k tweets, as the first in the community,
are collected and implemented as a dataset to evaluate the
performance of the proposed machine learning solution. These
tweets are collected with the related hashtags and keywords
of the Qatar World Cup 2022. The Vader algorithm is used
in this paper for sentiment analysis. Through the machine
learning method and collected Twitter tweets, we discovered
the sentiments and fun facts of several aspects important to
the period before the World Cup. The result shows people are
positive to the opening of the World Cup.

\end{abstract}

\begin{IEEEkeywords}
Sentiment Analysis, Social Media, Machine Learning
\end{IEEEkeywords}}

\maketitle

\IEEEdisplaynontitleabstractindextext

\IEEEpeerreviewmaketitle

\section{Introduction}\label{sec:introduction}
Social media is an important source for retrieving opinions from the public and provides interactions for different
users in the online world. Famous social media platforms,
like Twitter, Facebook, Youtube and Tiktok, consist of over
70
one of the popular international platforms for social media
and micro-blogging services, and the publicly posted microblogs are called tweets. Tweets allow users to connect
and update the topics and events related to their interests.
Although the tweets are usually short, the content can be
completed by different components like text content, emoji,
and hashtags. All these components provide sufficient information for opinion mining and sentiment analysis, and the
evaluation of these components can reflect the feelings or
polarity of the stand from the users. More than 400 million
tweets are posted every day on Twitter. Thus, a large dataset
can be collected to provide a convincing reflection of the
public attitude towards some special events. For example,
more than half a billion tweets have been obtained in the
previous World Cup.

With the large dataset and sentiment analysis, the public
reaction to the events during the World Cup period can be analyzed, providing information on public feelings and
attitudes. This paper investigates the relationship between
the sentiment based on Twitter tweet and the World Cup
2022 and contributes the following,


\begin{enumerate}
  \item Provide an open-source and updated dataset about FIFA 2022 with Twitter tweets;
  \item Analyze the excitement of the users about the World Cup 2022;
  \item Evaluate the popularity of football stars and football teams;
  \item Suggest an emoji handling solution for the tweet sentiment analysis.
\end{enumerate}
All the results are evaluated with the Vader algorithm, explanation and event spotting with the timeline. Over 350k tweets are investigated to provide a sufficient and convincing dataset from the public with the official hashtag of the FIFA World Cup 2022.

\section{Related Works}\label{sec:related_works}
Sentiment analysis is a natural language processing (NLP) technique that determines the opinion regarding the input text. There have been various methods to assess the sentiment of a text. Machine Learning methods have been implemented to classify them automatically. Both supervised and unsupervised learning methods have been shown to work well. Yue, Le et al.~\cite{10.1007/s10115-018-1236-4} have demonstrated multiple supervised and unsupervised learning techniques for sentiment analysis. Vader~\cite{Hutto2014VADERAP}is one of the unsupervised rule-based sentiment analysis models whose lexicon is specially tuned for social media. The Vader model works well on social media data and is efficient. It has the edge over its competition when used for Twitter data~\cite{DBLP:journals/corr/RibeiroAGBG15}. Vader algorithm also incorporates emoticons in its lexicon, which could help in better classification~\cite{Hutto2014VADERAP}.

There have been previous studies in identifying the sentiment of people during some crucial events  ~\cite{ElbagirAnalysisUN}~\cite{Barnaghi2015AnalysisAS}~\cite{Henk}. A similar study~\cite{Barnaghi2015AnalysisAS} was done for the 2014 FIFA World cup to identify people's sentiments during the event. They used manual labelling to train the Bayesian Logistic Regression model. 
Elbagir et al.~\cite{ElbagirAnalysisUN} have used the Vader model on Twitter data for the 2016 US elections. The result shows the model's effectiveness and efficiency. The multi-class classification helped them understand people's sentiments during the elections. 
Meier et al.~\cite{Henk} have shown the politicization of Mega sports. They have demonstrated how sentiments change before and after the FIFA World Cup 2018. Most topics, which were frequent before the World Cup, diminished as soon as the tournament started.

\subsection{Vader Algorithm}
Valence Aware Dictionary and sEntiment Reasoner (VADER) algorithm~\cite{Hutto2014VADERAP}~\cite{ElbagirAnalysisUN} is a lexicon and rule-based sentiment analysis tool that specifically attunes to sentiments expressed in social media. Besides, it provides a dictionary consisting of emoji sentiment, which other sentiment methods do not implement. A compound sentiment is computed based on the word and emoji in the tweets and normalized into the range $[-1, 1]$.
The sentiment, $S(x)$, for given input, $x$, is evaluated as,
\begin{equation} \label{eq:score}
    S(x) =  \sum_{i=1}^{N} \lambda_i \times f_i
\end{equation}
where $x:\{f_i|0<i\leq N\}$, $i$ is the feature index, $N$ is the total number of the extracted features, $\lambda_i$ is the sentiment of each lexicon, $f_i$ is the feature vector in each tweet.
\begin{table}[!h]
\caption{The value of $\lambda_i$ for different lexicons }
\begin{center}
\begin{tabular}{||c c||} 
 \hline
 Lexicon & $\lambda_i$ \\ [0.5ex] 
 \hline\hline
affected & -0.6\\ 
 \hline
\arm & 0.0\\
 \hline
loyalty & 2.5\\
\hline
\sunglass & 1.9\\
 \hline
\end{tabular}
\end{center}
\label{table:sentiment_sample}
\end{table}

Table~\ref{table:sentiment_sample} provides some examples of the sentiments evaluated from some words or emojis.
With the given sentiment based on the features, the compound sentiment, $S_c(x)$, can be evaluated as,
\begin{equation} \label{eq:compound_score}
    S_c(x) =  \frac{S(x)}{\sqrt{S(x)^2 + \alpha}}
\end{equation}
where $\alpha$ is the maximum expected value of the score, $15$, thus, the sentiment is bounded within $[-1, 1]$.
\begin{figure}[!t]
    \centering
    \includegraphics[width=\columnwidth]{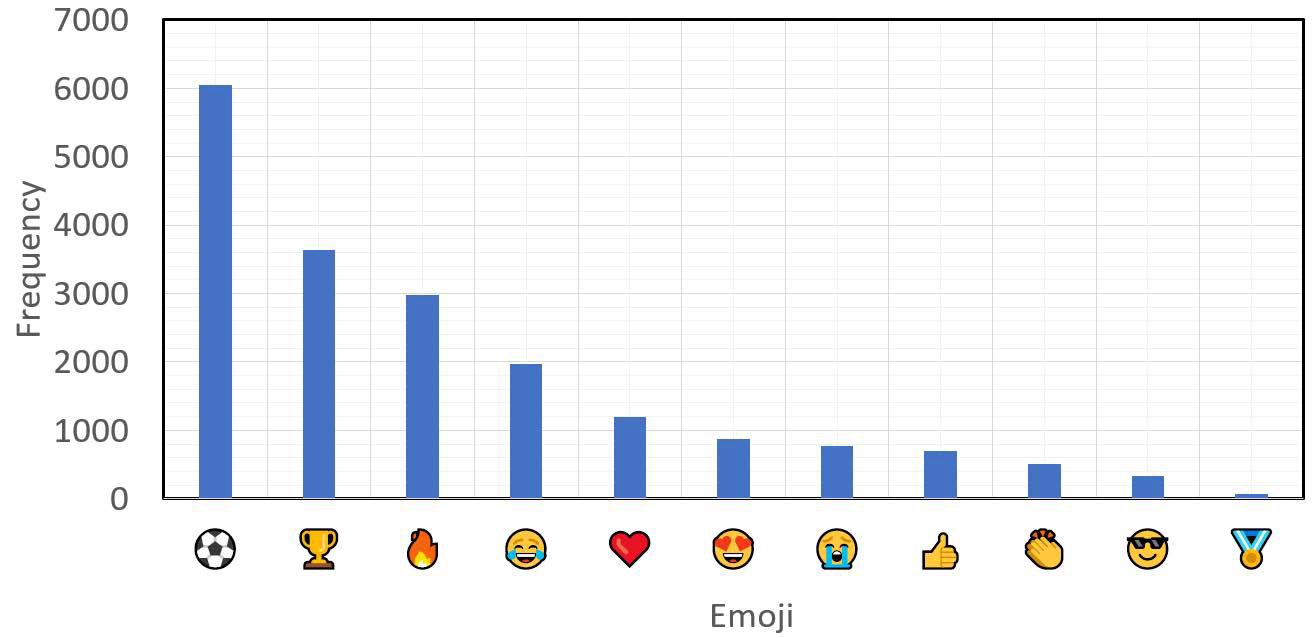}
    \caption{Emoji frequency in the collected tweets}
    \label{fig:emoji}
\end{figure}
Except for the features of text content, emojis also play an essential role in representing the users' feelings. Fig.~\ref{fig:emoji} shows the frequency of the emojis in the collected tweets, and around 30\% of tweets contain emojis. Therefore, sentiment analysis of emojis is essential to be considered.

\begin{table}[!h]
\caption{The influence of emoji in Vader algorithm}
\begin{center}
\begin{tabular}{||c c||} 
 \hline
 Tweets & Sentiment \\ [0.5ex] 
 \hline\hline
The FIFA is coming. I'm excited about it! & 0.4003\\ 
 \hline
The FIFA is coming. I'm excited about it! \arm& 0.4003\\
\hline
The FIFA is coming. I'm excited about it! \sunglass&0.69\\
 \hline
The FIFA is coming. I'm excited about it! \sunglass\sunglass &0.8268\\
\hline
\end{tabular}
\end{center}
\label{table:emoji_comparison}
\end{table}
Vader algorithm has special handling for the emojis in the tweets. The selection and the amount of emojis also affect the evaluated sentiment of the tweets which contain the same text content. Table 2 shows the addition or the amount of emojis may influence the sentiment. The emoji "\arm" provides a neutral polarity and the emoji "\sunglass" provides a positive polarity to the content. The increased number of "\sunglass" also makes the sentiment polarity more positive.
\begin{figure}[!t]
    \centering
    \includegraphics[width=0.8\columnwidth]{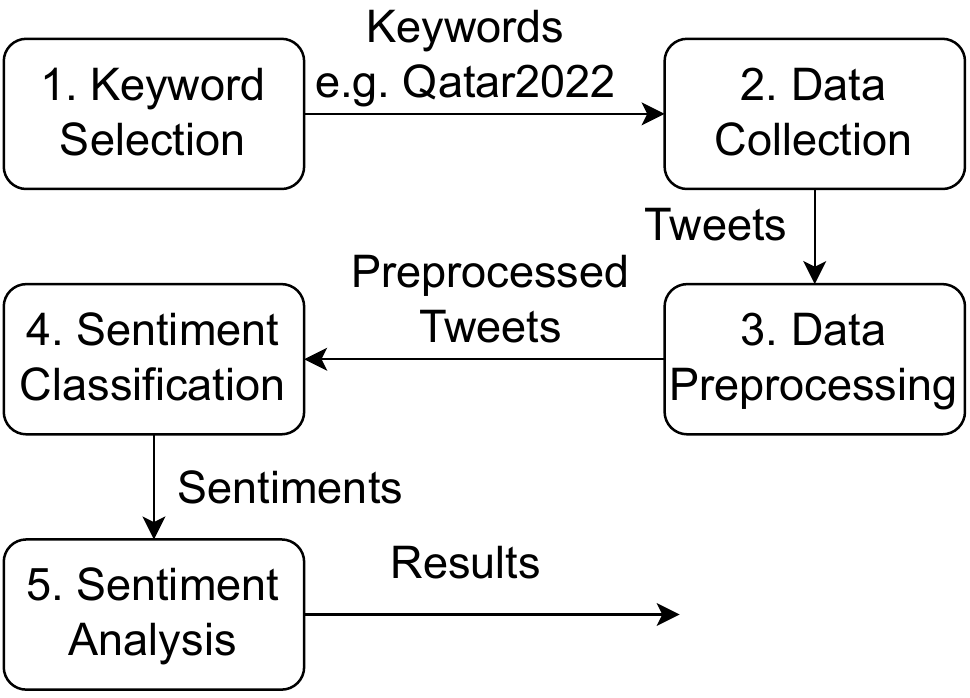}
    \caption{The flow and step of the sentiment analysis}
    \label{fig:flowchart}
\end{figure}

\section{Proposed Sentiment Analysis with Text and Emoji}
This section involved the flow, the proposed machine learning method and the details of each component in the sentiment analysis process. 

\subsection{The Flow and Step of the Sentiment Analysis}
The proposed solution involves multiple steps to retrieve the Twitter tweets and indicate their sentiment and polarity. The procedures are shown in Fig.~\ref{fig:flowchart} and described as,
\begin{enumerate}
  \item Keyword selection - select suitable keywords for the Twitter tweet collection;
  \item Data collection - collect and organize tweets by Twitter API with query parameters;
  \item Data pre-processing - remove the noise from the collected tweets;
  \item Sentiment classification - identify the polarity and calculate the sentiment of tweets;
  \item Sentiment analysis - evaluate the calculated polarity and sentiment of tweets.
\end{enumerate}
Throughout the whole process, the visualization of sentiment analysis is presented in Section~\ref{sec:result}.

\subsection{Selection of Keyword}\label{sec:keyword}
\begin{figure}[!t]
    \centering
    \includegraphics[width=\columnwidth]{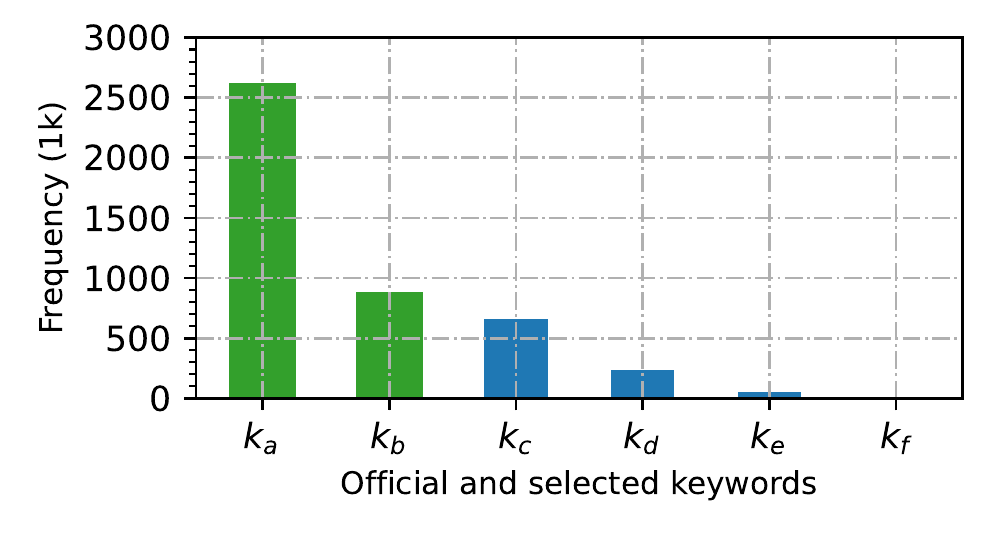}
    \caption{Frequency of keywords ($k_a$: \#FIFAWorldCup, $k_b$: \#Qatar2022, $k_c$: football world cup, $k_d$: qatar world cup, $k_e$: soccer world cup, $k_f$: worldcup2022), only $k_a$ and $k_b$ are official keywords}
    \label{fig:keyword}
\end{figure}
Keyword selection is an essential part that influences the
quality of the collected tweets for the sentiment analysis. In
this paper, official keywords and hashtags, i.e., Qatar2022
and FIFAWorldCup, are used for the crawling in the first
stage. However, the amount of collected tweets are small
based on these official hashtags, i.e., less than 2k. We
have also selected some other keywords and hashtags which
frequently exist in the tweets collected from the official
keywords, e.g., football world cup, worldcup2022, soccer
world cup and Qatar world cup. Fig. 3 shows the official
keywords occupy most of the frequency of keyword usage
in tweets and are followed by the suggested keywords based
on the tweet statistics. Thus, the selected keywords are the
most frequently used and related to the World Cup event.

\subsection{Data Collection}
After selecting the suitable keywords, we have used the
Twitter academic API for the data collection. It helps us
to retrieve all the tweets containing these keywords. Except
the keywords, different parameters can be added to the API
calls for the tweets needed based on the task requirement.
Therefore, we considered the filter of retweets to avoid
duplicated tweets or domination of certain tweet contents.
And we only collect the tweets in English everyday in the
period of 1 Aug., 2022 to 17 Sept., 2022. Around 130k
tweets are collected within the 48-day period

\subsection{Data Pre-processing}\label{sec:preprocess}
Pre-processing text is the first step in sentiment analysis. Proper methodologies are required to remove the noise from text without losing the essential semantic information. This important step involves dealing with hashtags, usernames, and emoticons. Our algorithm handles the emojis, and it has an impact on the sentiment. The unrelated tweets need to be filtered. Here is an example, 
\begin{displayquote}
Amazing giveaway i really excited \\@Ajay8307 @Tarun54552170 \\
\#NFTGiveaways \#NFTs \#Qatar2022
\end{displayquote}
The tweet is not related to the FIFA World Cup, but it used the official hashtag suggested by FIFA. If this tweet is used in the sentiment analysis, it will lead to a faulty result regarding the public feeling about World Cup 2022. The Twitter API has a parameter named exclude, which is used to filter out unwanted tweets. Here, by using \#NFTs, we can exclude the tweets with our keyword and the \#NFTs. 
Afterwards, the text of the tweet would be split using spaces. Each word would be used as a feature for our model.

\subsection{Sentiment Classification}
Given the collected tweets are pre-processed, all the tweets are taken into consideration of these classifications, (1) the sentiment; (2) the polarity; (3) the number of emoji that exists in the tweet.
The Vader algorithm evaluates the sentiment, which handles both the text and emoji content.
After the compound sentiment, $S_c(x)$, is evaluated for each tweet, the polarity is considered as positive if $S_c(x) > 0$, and it will be considered as negative if $S_c(x) < 0$. Otherwise, the remaining will be considered neutral. All data are organized in excel format and stored inside a server and Kaggle.

\subsection{Sentiment Result Analysis}
After the polarity and sentiment are stored on the server, we query each day's average sentiment and polarity within the collection period. Besides, the sentiment and polarity of all the tweets based on different keywords and selected football stars are evaluated. Based on the evaluation result, the observations are summarized in Section~\ref{sec:result}.

\section{Sentiment Finding and Fun Facts}\label{sec:result}
This section mentions the finding. The sentiments are compared versus time, and different football stars are investigated. The whole process is done within the period of World Cup 2022. Fun facts about the sentiment of tweets over time and about football stars are investigated.

\subsection{Sentiment analysis over time}
The sentiment is investigated versus the time during the FIFA World Cup period. The polarity of public feelings keeps positive within the 48 days before the opening of the World Cup.
\begin{figure}[!t]
    \centering
    \includegraphics[width=\columnwidth]{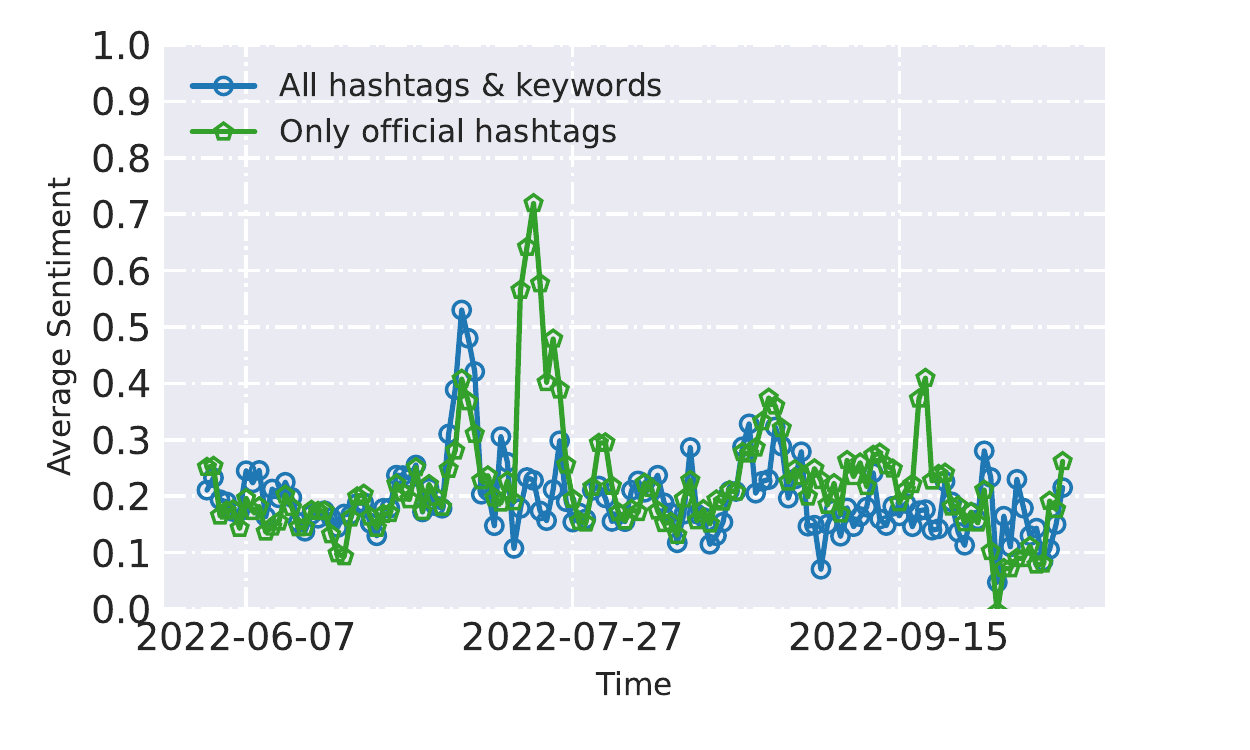}
    \caption{Sentiment analysis based on time}
    \label{fig:time_summary}
\end{figure}
Fig.~\ref{fig:time_summary} shows the average sentiment of the tweets related to the World Cup over time. All tweets that only use official keywords or all suggested keywords show the same trend of public expression in the World Cup.

\subsection{Sentiment analysis over football stars}

\begin{figure}[!t]
    \centering
    \includegraphics[width=\columnwidth]{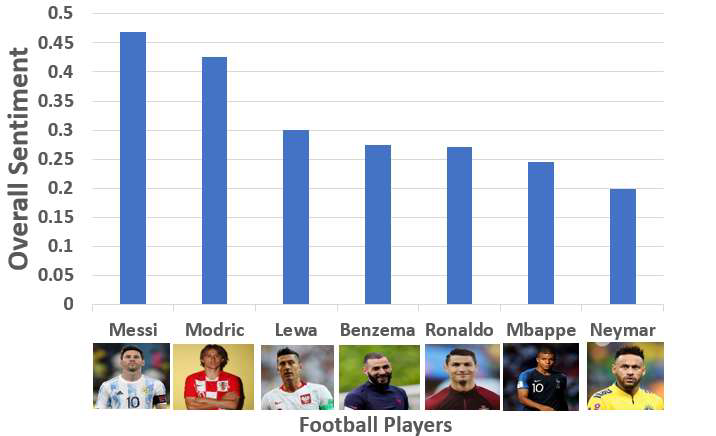}
    \caption{Sentiment analysis of tweets about football stars}
    \label{fig:player_summary}
\end{figure}

\begin{figure}[!t]
    \centering
    \includegraphics[width=\columnwidth]{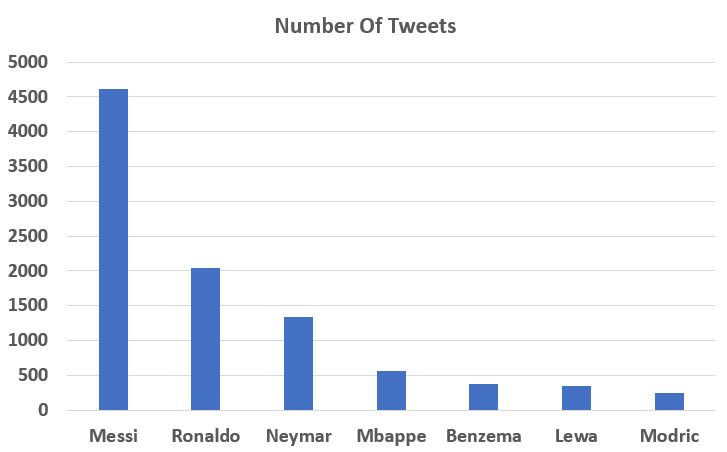}
    \caption{Number of tweets about football stars}
    \label{}
\end{figure}

Tweets could help us to reflect the public sentiment on
players’ on-field performance during the FIFA World Cup.
Different events concerning players or team members affect
how people think and talk about them. We are interested
in evaluating people’s sentiment toward the top players and
how it changes during the World Cup. The popularity of the
players may have some significance on how the players are
perceived by the general public.

\section{Future Work}
We would try to understand the public sentiment to help us to identify the golden ball winner in World Cup. Football players' sentiment must indicate their on-field performances, and a higher sentiment could help us to identify the winner. Some of the tweets consist of images, which are part of the feeling expressed by the public. Therefore, the image can provide extra information about the public's emotions. The sentiment correlation between text content and image can be investigated to enhance sentiment accuracy.

\section{Conclusion}
FIFA World Cup is a global event every four years, and Twitter, one of the most commonly used social platforms, provides tons of tweets to reflect the public's expression of these global events. With the data from Twitter, the public feelings can be analyzed based on the sentiment proposed solution with the collected tweets. And the main contributions can be summarized as follows,
\begin{enumerate}
  \item Provide an open-source and updated dataset about FIFA 2022;
  \item Provide a sentiment analysis that reflects the users are getting excited about the World Cup 2022 and proves that Twitter could be a social media sensor for the vibe of global users;
  \item Suggest an emoji handling solution for the tweet sentiment analysis.
\end{enumerate}
The sentiment accuracy can be further improved with a correlation study of the posted image and text content inside tweets instead of the pure emoji and text content study.

\ifCLASSOPTIONcompsoc
  \section*{Acknowledgments}
\else
  \section*{Acknowledgment}
\fi
This work was initiated by the College of Science and Engineering at HBKU and HKUST-NIE Social Media Lab. at Hong Kong University of Science \& Technology. The visualization and dataset storage is supported by CyPhy Media Limited.

\ifCLASSOPTIONcaptionsoff
  \newpage
\fi

\bibliographystyle{IEEEtran}
\bibliography{ref}

\clearpage

\appendices
\end{document}